\documentclass[10pt]{extarticle}
\usepackage[a4paper, total={6in, 9in}]{geometry}
\usepackage[utf8]{inputenc}
\usepackage{amsmath}
\usepackage{amsfonts}
\usepackage{amsthm}
\usepackage{amssymb}
\usepackage[table]{xcolor}
\usepackage{graphicx}
\usepackage[numbers, compress]{natbib}
\usepackage{hyperref}
\usepackage[inline]{enumitem}
\usepackage{wrapfig}
\usepackage{caption}
\usepackage{subcaption}
\usepackage{xspace}
\usepackage{booktabs}
\usepackage{multirow}
\usepackage[inline]{enumitem}

\usepackage[utf8]{inputenc} 
\usepackage[T1]{fontenc}    
\usepackage{hyperref}       
\usepackage{url}            
\usepackage{booktabs}       
\usepackage{amsfonts}       
\usepackage{nicefrac}       
\usepackage{microtype}      
\usepackage{xcolor}         

\usepackage{amsmath}
\usepackage{amsfonts}
\usepackage{amsthm}
\usepackage{amssymb}
\usepackage{graphicx}
\usepackage{hyperref}
\usepackage{wrapfig}
\usepackage{caption}
\usepackage{subcaption}
\usepackage{xspace}
\usepackage{multirow}
\usepackage[inline]{enumitem}
\usepackage{cases}

\newtheorem{assumption}{Assumption}

\newcommand{\R}{\mathbb{R}}
\newcommand{\E}{\mathbb{E}}
\newcommand{\cov}{\mathrm{Cov}}
\newcommand{\vect}[1]{\boldsymbol{\mathbf{#1}}}
\newcommand{\eqdef}{\doteq}

\newcommand{\x}{\vect{x}}
\newcommand{\h}{\vect{h}}
\newcommand{\y}{\vect{y}}
\newcommand{\Vx}{\mathbb{V}_{\x}}
\newcommand{\Vh}{\mathbb{V}_{\h}}
\newcommand{\Vy}{\mathbb{V}_{\y}}

\newcommand{\s}{\vect{s}}
\newcommand{\z}{\vect{z}}

\newcommand{\mF}{\vect{F}}
\newcommand{\mP}{\vect{P}}
\newcommand{\mA}{\vect{A}}
\newcommand{\mW}{\vect{W}}
\newcommand{\veta}{\vect{\eta}}
\newcommand{\vnu}{\vect{\nu}}
\newcommand{\vetaA}{\vect{\alpha}}

\title{\bfseries Graph Kalman Filters}

\date{}

\usepackage{authblk}
\author[12]{Cesare Alippi\footnote{Equal contribution.}}
\newcommand\CoAuthorMark{\footnotemark[\arabic{footnote}]}
\author[1]{Daniele Zambon\protect\CoAuthorMark\thanks{Corresponding author: \texttt{daniele.zambon@usi.ch}}}

\affil[1]{\small The Swiss AI Lab IDSIA \& Universit\`a della Svizzera italiana, Switzerland.}
\affil[2]{\small Politecnico di Milano, Italy.}

\begin{document}

\maketitle

\begin{abstract}
The well-known Kalman filters model dynamical systems by relying on state-space representations with the next state updated, and its uncertainty controlled, by fresh information associated with newly observed system outputs. This paper generalizes, for the first time in the literature, Kalman and extended Kalman filters to discrete-time settings where inputs, states, and outputs are represented as attributed graphs whose topology and attributes can change with time. The setup allows us to adapt the framework to cases where the output is a vector or a scalar too (node/graph level tasks). Within the proposed theoretical framework, the unknown state transition and readout are learned end-to-end along with the downstream prediction task.
\end{abstract}


\section{Introduction}

The Kalman Filter (KF) \citep{kalman1960new} is a state-space representation architecture for modeling dynamical systems. Since its introduction more than 60 years ago, the KF has been standing out for its performance in tracking and controlling applications as well as for its relative simplicity. 
The KF operates on linear dynamical systems of the form
\begin{equation}\label{eq:state-space-linear-time-inv} 
\begin{cases}
\h_t = \mF\, \h_{t-1} + \vect G\, \x_{t-1} + \veta_{t-1}, 
    \\
\y_t = \vect H\, \h_{t} + \vect \vnu_{t},
\end{cases}
\end{equation}
by estimating the hidden system state vector $\h_t\in \R^{d_h}$ at time $t$ given input vector $\x_{t-1}\in\R^{d_x}$ and output vector $\y_t\in\R^{d_y}$; $\{\veta_t\}$ and $\{\vnu_t\}$ are Gaussian white-noise stochastic processes. 
At each time step $t$, the system state and its uncertainty update by iteratively leveraging on previous estimates and incorporating newly observed system outputs and inputs while accounting for uncertainties. Notably, KF estimators are proven to be optimal, being unbiased and of minimum variance.
Generalizations and variants of the KF cover continuous-time reformulations of \eqref{eq:state-space-linear-time-inv}, non-Gaussian noise distributions, time-variant state-space setups, and nonlinear formulations of \eqref{eq:state-space-linear-time-inv} that can be cast in the form
\begin{equation}\label{eq:state-space-nonlinear}
\begin{cases}
\h_t = f_\textsc{st}(\h_{t-1},\x_{t-1}) + \veta_{t-1},
\\
\y_t = f_\textsc{ro}(\h_t) + \vnu_t;
\end{cases}
\end{equation}
we refer the reader to \cite{simon2006optimal} and Section~\ref{sec:related-work} for a review.
Nowadays graph-based models integrating relational information among the sensors/components of multivariate systems have been demonstrated to be extremely powerful and effective spatio-temporal predictors \cite{li2018diffusion,wu2019graph,hansen2022power}. These include graph neural networks \cite{bacciu2020gentle,bronstein2017geometric} and their extensions incorporating temporal information, usually referred to as spatio-temporal graph neural networks (STGNNs) \cite{seo2018structured,ruiz2020gated,gao2022equivalence}.
The literature on KF with graph-structured data is however less mature, as discussed in Section~\ref{sec:related-work}.

\paragraph{Contribution}
In this paper, for the first time in the literature, we provide a graph-based version of the KF where inputs,  outputs, and states are attributed graphs whose topology is allowed to change over time. Its neural deep nonlinear implementation derives from a graph state space (GSS) \citep{zambon2023graph} modeling a discrete-time, time-invariant, stochastic data-generating process $\mathcal P$
\begin{equation}\label{eq:state-space}
\begin{cases}
\h_t = f_\textsc{st}(\h_{t-1},\x_{t-1}, \veta_{t-1}),
\\
\y_t = f_\textsc{ro}(\h_t, \vnu_t),
\end{cases}
\end{equation}
where inputs $\x_{t-1}$, states $\h_t$, and outputs $\y_t$ are attributed graphs belonging to graph spaces $\mathcal X$, $\mathcal H$, and  $\mathcal Y$, respectively. 
Stochastic processes $\{\veta_t\}$, $\{\vnu_t\}$ are white noise impacting on the node signals and/or the topology of the graphs.

The GSS formulation \eqref{eq:state-space} poses three main challenges addressed in this paper.
Firstly, functions $f_\textsc{st}$ and $f_\textsc{ro}$ are assumed to be unknown and, differently from \eqref{eq:state-space-nonlinear}, nonlinear also with respect to the noise components.
Secondly, unknown states are attributed graphs of unknown and time-varying topology, hence it is requested us to estimate both node features and graph topology. Note that the topology is a discrete entity;  as such, it is not amenable to standard gradient-based optimization.
Lastly, the data dimensionality -- say the nodes and edges of the involved graphs -- is not fixed and can be very large, yielding ill-posed estimation, in general. 
We address the above challenges, by devising a spatio-temporal graph neural network (STGNN) that approximates the state-transition and the readout functions. 

After reviewing the standard KF and one of its generalizations to nonlinear vector-based systems in Section~\ref{sec:KF}, we introduce the considered GSS system model  
and present a parametric family of GSS models to learn the system dynamics in Section~\ref{sec:GSS}. In Section~\ref{sec:GKF}, we derive the proposed Graph KF architecture and validate it in Section~\ref{sec:experiments}.

\section{Kalman filters}\label{sec:KF}

Consider the discrete time-invariant system model \eqref{eq:state-space-nonlinear} with random initial state $\h_0\in\R^{d_h}$ drawn from a known finite-variance distribution, \emph{state transition} $f_\textsc{sc}$ and \emph{readout} $f_\textsc{ro}$ are differentiable with respect to the states and affected by white-noise stochastic processes with covariance matrices $\cov[\veta_t]=\vect Q_t$ and $\cov[\vnu_t]=\vect R_t$, for all $t$. 
Let $\h_0,\veta_t$ and $\vnu_t$ be mutually independent, for all $t$.
 
Assume to have observed $\x_{i-1},\y_i$ for all $i<t$ and generated an estimate $\h_{t-1}^+$ of $\E[\h_{t-1}]$ with error covarinace matrix $P_{t-1}^+\eqdef \cov[\h_{t-1}-\h_{t-1}^+]$.
A single iteration of the KF algorithm aimed at modeling \eqref{eq:state-space-nonlinear} is two-step: 
\begin{enumerate*}[label={(\roman*)}]
    \item 
    Once input $\x_{t-1}$ is available, an \emph{a priori} estimate $\h_t^-$ of $\E[\h_t]$ is produced along with error covariance matrix $\mP_{t}^-\eqdef\cov[\h_t-\h_t^-]$ and followed by a prediction $\y_t^-=f_\textsc{st}(\h_t^-)$ of the system output $\y_t$; estimates denoted with superscript ``$-$'' are named ``a priori'' as they are obtained before observing the system output $\y_t$. 
    \item 
    Once $\y_t$ is observed, an \emph{a posteriori} estimate $\h_t^+$ refines $\h_t^-$ and the updated matrix $\mP_{t}^+\eqdef \cov[\h_t-\h_t^+]$ is derived from $\mP_{t}^-$.
\end{enumerate*}

\subsection{KF for linear systems}
\label{sec:KF-lin}

This section derives the KF procedure for discrete-time, time-variant, linear systems
\begin{equation}\label{eq:state-space-linear-time-var} 
\begin{cases}
\h_t = \mF_{t-1} \h_{t-1} + \vect G_{t-1} \x_{t-1} + \veta_{t-1}, 
    \\
\y_t = \vect H_{t} \h_{t} + \vnu_{t},
\end{cases}
\end{equation}
a generalization of the time-invariant system \eqref{eq:state-space-linear-time-inv}
where matrices $\mF_{t-1}$, $\vect G_{t-1}$ and $\vect H_t$ depends on $t$.
Although we aim at developing a KF for time-invariant GSS models like \eqref{eq:state-space}, in order to deal with nonlinear state transition and readout,
it is suitable to rely on
the following time-variant derivation.

\paragraph{A priori estimate} Note that
$\E[\h_t]= 
\mF_{t-1}\E[\h_{t-1}] + \vect G_{t-1}\x_{t-1} + 0$,
so, if $\E[\h_{t-1}^+]=\E[\h_{t-1}]$, then the following is an unbiased estimator of $\h_t$:
\begin{equation}\label{eq:kf:ht-}
    \h_t^- \eqdef \mF_{t-1} \h_{t-1}^+ + \vect G_{t-1}\x_{t-1};
\end{equation}
assume $\h_0^+=\E[\h_0]$. The covariance matrix of the a priori estimation error is
\begin{align}
    \mP_t^- &= \cov[\h_t - \h_t^-] = \E[(\h_t - \h_t^-)(\h_t - \h_t^-)^\top],
\end{align}
as $\E[\h_{t}^-]=\E[\h_{t}]$, and is expressed as function of the $\mP_{t-1}^+$, from the previous time step:
\begin{align}
\mP_t^-
        &=\cov[\mF_{t-1}(\h_{t-1}-\h_{t-1}^+)] + \cov[\veta_{t-1}]
\label{eq:kf:Pt-}
        =\mF_{t-1}\mP_{t-1}^+\mF_{t-1}^\top + \vect Q_{t-1},
\end{align}
given the independence between $\veta_{t-1}$ and both $\h_{t-1}$ and $\h_{t-1}^+$.

\paragraph{A posteriori estimate}
The a posteriori estimate has the form
\begin{align}\label{eq:kf:ht+}
    \h_t^+ &\eqdef \h_t^- + \vect K_t (\y_t - \y_t^-)
\end{align}
where matrix $\vect K_t$ is known as \emph{gain} while residual $\y_t - \y_t^-$ is called \emph{innovation}. As $\h_t^-$ is unbiased, then $\E[\y_t-\y_t^-]=0$, and we see that $\h_t^+$ is unbiased as well, regardless of the choice of the gain. Therefore, we can select $\vect K_t$ to minimize the total variance
\begin{equation}
    \textrm{Tr}(\mP_t^+)=\E[(\h_t-\h_t^+)^\top(\h_t-\h_t^+)],
\end{equation}
i.e., the trace of the matrix $\mP_t^+$.
By exploiting the independence between $\h_t^-$ and $\vnu_t$, we get
\begin{align}
    \h_t^+ &= \h_t^- + \vect K_t (\vect H_t(\h_t -\h_t^-)+\vnu_t)
\\
    \h_t - \h_t^+ 
    &= (\h_t -\h_t^-) - \vect K_t \vect H_t(\h_t -\h_t^-) - \vect K_t\vnu_t
    = (\mathbb I - \vect K_t \vect H_t)( \h_t - \h_t^-) - \vect K_t \vnu_t
\\\label{eq:kf:Pt+}
    \mP_t^+ &= (\mathbb I - \vect K_t \vect H_t)\mP_t^-(\mathbb I - \vect K_t \vect H_t)^\top + \vect K_t \vect R_t\vect K_t^\top.
\end{align}
Finally, the gain minimizing $\textrm{Tr}(\mP_t^+)$ is
\begin{equation}
    \label{eq:kf:Kt}
    \vect K_t = \mP_t^-\vect H_t^\top (\vect H_t\mP_t^-\vect H_t^\top + \vect R_t)^{-1};
\end{equation}
for further developments see, e.g., \cite{simon2006optimal}. 

\paragraph{KF steps} Given the above, the KF iteration for the generic time step $t$ is
\begin{align}
    \h_t^- &\stackrel{\eqref{eq:kf:ht-}}= \mF_{t-1} \h_{t-1}^+ + \vect G_{t-1}\x_{t-1}
    &
    \mP_t^- &\stackrel{\eqref{eq:kf:Pt-}}= \mF_{t-1}\mP_{t-1}^+\mF_{t-1}^\top + \vect R_t
    \\
    \y_t^- &\stackrel{\eqref{eq:state-space-linear-time-var}}= \vect H_t \h_t^-
    &
    \vect K_t &\stackrel{\eqref{eq:kf:Kt}}= \mP_t^-\vect H_t^\top (\vect H_t\mP_t^-\vect H_t^\top + \vect R_t)^{-1}
    \\
    \h_t^+ &\stackrel{\eqref{eq:kf:ht+}}= \h_t^- + \vect K_t (\y_t - \y_t^-)
    &
    \mP_t^+ &\stackrel{\eqref{eq:kf:Pt+}}= (\mathbb I - \vect K_t \vect H_t)\mP_t^-(\mathbb I - \vect K_t \vect H_t)^\top + \vect K_t \vect R_t\vect K_t^\top
\end{align}

Finally, we mention that the literature shows different equivalent rewritings to meet specific implementational requirements \citep{simon2006optimal}. Such derivations are out of this paper's scope.

\subsection{Extended KF for nonlinear systems}\label{sec:EKF}

The Extended KF (EKF) \cite{smith1962application} adapts the KF of Section~\ref{sec:KF-lin} to the nonlinear case of \eqref{eq:state-space-nonlinear}. EKF operates by linearizing the state-transition and readout functions around the last available state estimate.
EKF requires the first-order Taylor approximation of function $f_\textsc{st}(\,\cdot\,, \x_{t-1})$ around $\h_{t-1}^+$
\begin{align}\label{eq:ekf:Ft}
    f_\textsc{st}(\h, \x_{t-1})
&\approx f_\textsc{st}\left( \h_{t-1}^+, \x_{t-1}\right) 
    + 
    \underbrace{\nabla_{\h} f_\textsc{st}(\h_{t-1}^+, \x_{t-1})}_{\mF_{t-1}}
    (\h - \h_{t-1}^+) 
\\\label{eq:ekf:xt-tilde}
&= 
    \mF_{t-1}\h  + f_\textsc{st}\left( \h_{t-1}^+, \x_{t-1}\right) - \mF_{t-1}\h_{t-1}^+.
\end{align}
Similarly, we expand with Taylor function $f_\textsc{ro}$ around estimate $\h_{t}^-$
\begin{align}\label{eq:ekf:Ht}
    f_\textsc{ro}(\h)
&\approx f_\textsc{ro}\left( \h_{t}^-\right) 
    + \big(\underbrace{\nabla_{\h} f_\textsc{ro}(\h_{t}^-)}_{\vect H_t}\big) (\h - \h_{t}^-) 
= 
    \vect H_{t}\h  + f_\textsc{ro}\left( \h_{t}^-\right) - \vect H_{t}\h_{t}^-.
\end{align}
Derivations lead to the following computation at every time instant $t$:
\begin{align}
    \mF_{t-1} &\stackrel{\eqref{eq:ekf:Ft}}=\nabla_{\h} f_\textsc{st}(\h_{t-1}^+, \x_{t-1})
    &
    \vect H_t &\stackrel{\eqref{eq:ekf:Ft}}= \nabla_{\h} f_\textsc{ro}(\h_{t}^-)
    \\
    \h_t^- &\stackrel{\eqref{eq:ekf:xt-tilde}}=
    f_\textsc{st}\left( \h_{t-1}^+, \x_{t-1}\right)
    &
    \mP_t^- &\stackrel{\eqref{eq:kf:Pt-}}= \mF_{t-1}\mP_{t-1}^+\mF_{t-1}^\top + \vect R_t
    \\
    \y_t^- &\stackrel{\eqref{eq:state-space-nonlinear}}= 
    f_\textsc{ro}\left( \h_{t}^-\right)
    &
    \vect K_t &\stackrel{\eqref{eq:kf:Kt}}= \mP_t^-\vect H_t^\top (\vect H_t\mP_t^-\vect H_t^\top + \vect R_t)^{-1}
    \\
    \h_t^+ &\stackrel{\eqref{eq:kf:ht+}}= \h_t^- + \vect K_t (\y_t - \y_t^-)
    &
    \mP_t^+ &\stackrel{\eqref{eq:kf:Pt+}}= (\mathbb I - \vect K_t \vect H_t)\mP_t^-(\mathbb I - \vect K_t \vect H_t)^\top + \vect K_t \vect R_t\vect K_t^\top.
\end{align}

As anticipated at the beginning of Section~\ref{sec:KF}, by linearizing the time-invariant system \eqref{eq:state-space-nonlinear} we obtain a linear time-variant system like \eqref{eq:state-space-linear-time-var}, where $\mF_{t-1}$, $\vect G_{t-1}$, and $\vect H_t$ depend on the given input and current state estimates.
The EKF is applicable to more general system models, where the interaction with the noise processes is nonlinear. We expand the discussion in Section~\ref{sec:GKF}, when deriving the KF for graphs.

\section{Related work}
\label{sec:related-work}

The EKF \cite{smith1962application} has been further generalized to account for orders beyond the first \cite{athans1968suboptimal}. The unscented KF employs particle filtering to address some of the drawbacks of EKF \cite{julier2004unscented,menegaz2015systematization}. The theory of reproducing kernel Hilbert space is another viable solution to operate with nonlinear systems and non-Gaussian noise \cite{ralaivola2005time,chen2017maximum,dang2019kernel}. 
Regarding graph data, the research focused on linear systems with known topology \cite{shi2009kalman,isufi2019forecasting,isufi2020observing}. With \cite{romero2017kernelbased} introducing the analysis over a known, dynamic topology.

\section{Graph state-space models}
\label{sec:GSS}

\begin{figure}
\centering
\includegraphics[width=.7\textwidth]{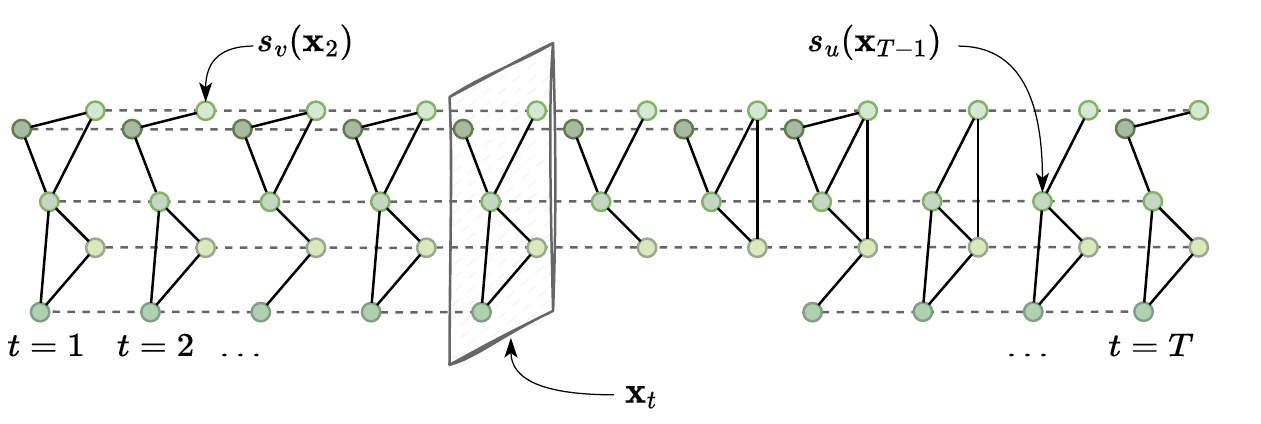}
\caption{An example of a spatio-temporal data over a set $\Vx$ of 5 nodes. Graph $\x_{t-1}$ is given at each step $t$. $\x_{t-1}$ is defined over a node set $V(\x_{t-1})\subseteq\Vx$ and has node signals $s_v(\x_{t-1})\in\mathbb R^{d_x}$ associated with each given node.}
\label{fig:dynamic-graph}
\end{figure}

The input graph $\x_t\in\mathcal X$ at time $t$ is defined over node set $V(\x_t)$, e.g., associated with the sensors of a sensor network, and edge set represented as adjacency matrix $\mA(\x_t)$ that encodes the relations existing among the nodes, such as physical proximity, signal correlations, or causal dependencies. 
The node sets and the particular topologies observed at different time steps $t,t'$ are generally different but, typically, $V(\x_t)\cap V(\x_{t'})\ne \emptyset$, implying the existence of a partial correspondence among groups of nodes. We denote with $\Vx = \bigcup_{t} V(\x_t)$ the union set of all nodes, whose cardinality is assumed to be finite. 
Input graphs are attributed with node features attached to them, like sensor readings, collected in graph signal $s(\x_t)\in\R^{|\Vx|\times d_x}$. 
A visual representation of the resulting graph-based spatio-temporal data sequence $\x_1,\x_2,\dots,\x_t,\dots$ is provided in Figure~\ref{fig:dynamic-graph}.

Given sequence $\{\x_t\}_t$ of graphs defined over node set $\Vx$, we aim at predicting output graphs $\y_t\in\mathcal Y$ at each time step $t$. 
We model the data-generating process $\mathcal P$ as formulated in \eqref{eq:state-space}, with system model
\begin{equation}\label{eq:state-space-system-model}
\begin{cases}
\h_t = f_\textsc{st}(\h_{t-1},\x_{t-1}, \veta_{t-1}),
\\
\y_t = f_\textsc{ro}(\h_t, \vnu_t),
\end{cases}
\end{equation}
involving representations of the system states as graphs $\h_t \in \mathcal H$; 
with a consistent notation, $V(\h_t)\in\Vh, \mA(\h_t)$, and $s(\h_t)$ ($V(\y_t)\in\Vy, \mA(\y_t)$, and $s(\y_t)$) denote the node set, the adjacency matrix and the signal of state graph $\h_t$ (output graph $\y_t$).
Stochastic processes $\{\veta_t\}$ and $\{\vnu_t\}$ are white noises impacting the edges and the node signals of the states \cite{zambon2022aztest}.
Functions $f_\textsc{st}$ and $f_\textsc{ro}$, as well as the noise distributions, are assumed unknown with finite second moments.
Finally, exogenous variables, like those referring to extra sensor information or functional relations, can be included as well in the framework and encoded in $\x_t$ to avoid overwhelming notation.
As better discussed in \cite{zambon2023graph}, the GSS \eqref{eq:state-space-system-model} does not require any identification between the nodes in sets $\Vx$, $\Vh$, and $\Vy$, although this might be the case in some scenarios.

\begin{figure}
    \vspace{0.5cm}
    \centering
    \includegraphics[width=\textwidth]{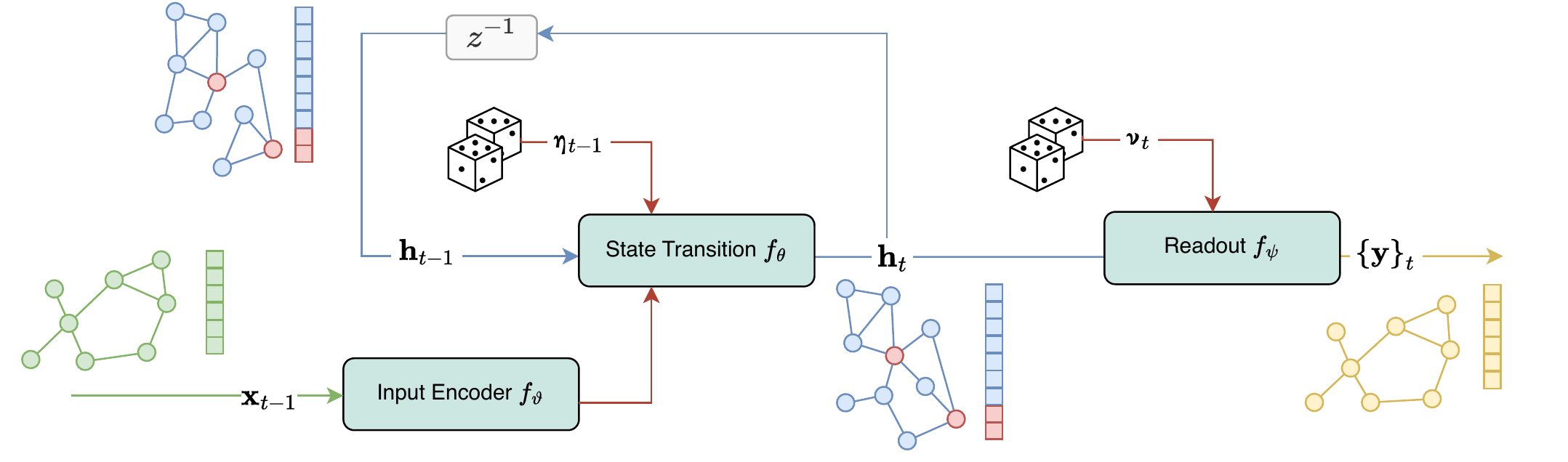}
    \caption{Block diagram of the GSS stochastic model of \eqref{eq:state-space-approx-generic}.}
    \label{fig:gss-approx}
\end{figure}

We introduce a GSS family of stochastic predictive models \citep{zambon2023graph}
\begin{equation}\label{eq:state-space-approx-generic}
\begin{cases}
\h_t = f_{\theta}(\h_{t-1}, f_\vartheta(\x_{t-1}), \veta_{t-1})
\\
 \y_t = f_\psi(s(\h_t), \vnu_t)
\end{cases}
\end{equation}
where $\h_t,\y_t,\veta_{t-1}$ and $\vnu_t$ are random variables. Parameters $\theta,\varphi$, and $\psi$ are learned from data, i.e., from a realization of process $\mathcal P$. 
Initial condition $\h_0$ is drawn from a given prior distribution $P_{\h_0}$. 
Function $f_\vartheta$ is the input encoder, mapping $\x_{t-1}$ to the nodes of graph space $\mathcal H$, function $f_\theta$ models the state transition inferring both the graph topology of $\h_t$ and the associated node signal, whereas $f_\psi$ is the readout. 
Graph functions $f_\vartheta$, $f_\theta$, and $f_\psi$ are parametrized in real vectors $\vartheta,\theta,$ and $\psi$, respectively, and learned directly from data; we assume them differentiable with respect to the associated parameter vectors.
A schematic view of \eqref{eq:state-space-approx-generic} is given in Figure~\ref{fig:gss-approx}.

Importantly, and differently from what is typically done with state-space models, state representations are first predicted and then refined once the system output is observed, in line with traditional vectorial KF. While training the model parameters is carried out with standard deep learning techniques, the state estimate refinement follows the KF proposed here and is derived in the following section.

\section{Graph Kalman filter}\label{sec:GKF}

The proposed Graph KF follows the linearization of the EKF, taking care of differentiating  with respect to the noise components, too. 
Accordingly, we assume that both $f_\theta$ and $f_\psi$ are differentiable with respect to the state and the noise terms, as better formalized in following Section~\ref{sec:GKF:linearize}.
To facilitate readability, we assume the following.
\begin{assumption}\label{a:ht-nodes} 
The node sets of input, state, and output graphs coincide ($\Vy=\Vh=\Vx$).
\end{assumption}
\begin{assumption}\label{a:fixed-topology} 
The adjacency matrix $\mA(\h_t)$ is written as 
$\mA(\h_t)=\mA_\theta + \vetaA_{t-1}$,
where $\vetaA_{t-1}$ is a random ${|\Vh|\times |\Vh|}$ matrix accounting for the randomness associated with the state-transition function $f_\textsc{st}$ (i.e., $\veta_{t-1}=\vetaA_{t-1}$). 
\end{assumption}
\begin{assumption}\label{a:yt-topology} 
The topology of the output graph is either that of the state ($\mA(\y_t)=\mA(\h_t)$) or the input ($\mA(\y_t)=\mA(\x_{t-1})$).
\end{assumption}

We stress that above assumptions are either made to make the derivation more amenable or ease the readability of the outcomes by providing a simplified notation. 
For instance, Assumption~\ref{a:ht-nodes} ensures an immediate correspondence between nodes of the inputs, states, and outputs (without the need for graph pooling and upscaling operators as in \citep{zambon2023graph}), whereas predicting the output adjacency matrix $\mA(\y_t)$ -- thus relaxing Assumption~\ref{a:yt-topology} -- would only request additional components of $f_\psi$ to be linearized.
It follows that weaker assumptions can be considered at the cost of more complex mathematics that, however, will not change the spirit of what is derived.

\subsection{State-transition and readout functions linearization}\label{sec:GKF:linearize}

Note that, by Assumption \ref{a:fixed-topology}, $\mA_\theta$ is encoded by parameter vector $\theta$ and can be incorporated in $f_\theta$ so that GSS model \eqref{eq:state-space-approx-generic} can be simplified to
\begin{equation}\label{eq:state-space-approx-assumptions}
\begin{cases}
\h_t = f_{\theta}(s(\h_{t-1}), f_\vartheta(\x_{t-1}), \vetaA_{t-1})
\\
 \y_t = f_\psi(s(\h_t), \vnu_t);
\end{cases}
\end{equation}
in the following, denote signal $s(\h_t)$ as $\s_t$, for brevity. 
We note that, although the noise term $\vetaA_{t-1}$ is added to $\mA_\theta$ to form $\mA(\h_t)$ (Assumption~\ref{a:fixed-topology}), $\vetaA_{t-1}$ impacts on the signal $s(\h_t)$ too if $f_\theta$ is, e.g., an STGNN.
Moreover, with \eqref{eq:state-space-approx-assumptions} we requests $f_\theta$ to be differentiable with respect to $\s_{t-1}$ and $\vetaA_{t-1}$, representable as a vector in $\R^{|\Vh|}$ and a matrix in $\R^{|\Vh|\times|\Vh|}$, respectively; similarly, $f_\psi$ to be differentiable with respect to $\s_t$ and $\vnu_t$.

Following the EKF procedure, we expand the nonlinear system \eqref{eq:state-space-approx-assumptions} with Taylor approximating $f_\theta(\s_{t-1}^+f_\vartheta(\x_{t-1}),\vetaA_{t-1})$ and $f_\psi(\s_t^-,\vnu_t)$.
\begin{align}
f_\theta(\s, \tilde\x_{t-1},\vetaA)
&\approx f_\theta\left( \s_{t-1}^+, \tilde\x_{t-1}, \vect 0 \right)
    + \left(\nabla_{\s} f_\theta({\s_{t-1}^+, \tilde\x_{t-1}, \vect 0})\right) (\s - \s_{t-1}^+) 
    \\&\qquad\qquad\qquad\qquad\qquad\qquad\qquad\qquad+ \left(\nabla_{\vetaA} f_\theta({\s_{t-1}^+, \tilde\x_{t-1}, \vect 0})\right)\bullet \vetaA
\\&= f_\theta\left( \s_{t-1}^+, \tilde\x_{t-1}, \vect 0 \right) 
    +  \mF_{t-1} (\s - \s_{t-1}^+) 
    + \vect L_{t-1}\bullet  \vetaA
\\&
= \mF_{t-1} \s 
    + {f_\theta\left( \s_{t-1}^+, \tilde\x_{t-1}, \vect 0 \right) -  \mF_{t-1} \s_{t-1}^+}
    + {\vect L_{t-1} \bullet \vetaA}
\end{align}
where  $\tilde\x_{t-1}=f_\varphi(\x_{t-1})$ and $\vect B\bullet \vect C\in \R^{|\Vh|}$ denotes the product 
\begin{equation}\
    [\vect B\bullet \vect C]_v={\sum}_{i,j=1}^{|\Vh|} \vect B_{v,i,j}\vect C_{i,j}
\end{equation}
for all $\vect B\in\R^{|\Vh|\times|\Vh|\times|\Vh|}$ and $\vect C\in\R^{|\Vh|\times|\Vh|}$. Similarly, we linearize the readout function:
\begin{align}
f_\psi(\s, \vnu) &\approx f_\psi(\s_t^-,\vect 0) 
    + (\underbrace{\nabla_{\s} f_\psi({\s_t^-,\vect 0})}_{\vect H_t}) (\s-\s_t^-)
    + (\underbrace{\nabla_{\vnu} f_\psi({\s_t^-,\vect 0})}_{\vect M_t}) {\,\vnu}
    \\ &= 
    \vect H_t \s
    + {f_\psi(\s_t^-,\vect 0) - \vect H_t \s_t^-}
    + {\vect M_t\vnu_t}.
\end{align}
We stress that, even though the graph topology might not be visible in the above notation, graph-based processing is still carried out within the linear operators defined by matrices $\mF_{t-1}$ and $\vect L_{t-1}$.

\subsection{Graph KF iterations}
\label{sec:GKF-inference}

Assume to have learned parameter vectors $\vartheta,\theta,$ and $\psi$ of GSS model \eqref{eq:state-space-approx-assumptions}, then the following iterations define the proposed Graph KF.
Initialize 
\begin{align}
    \s_0^+ &= \E_{\s\sim P_{\s_0}}[\s] \in \R^{|\Vh|}
&
    \vect P_0^+ &= \cov_{\s\sim P_{\s_0}}[\s] \in \R^{|\Vh|\times|\Vh|},
\end{align}
from a prior distribution $P_{\s_0}$. 
Then, for $t=1,2,3,\dots$ 
\begin{enumerate}[label=(\roman*)]

    \item Encode input graph: 
    \label{step:gkf:xt-enc}
    \begin{align}
        \label{eq:gkf:xt-enc}
        \tilde\x_{t-1} &= f_\vartheta(\x_{t-1})
    \end{align}

    \item Update the a priori state estimate:
    \label{step:gkf:st-} 
    \begin{align}
        \label{eq:gkf:st-}
        \s_t^- &= f_\theta(\s_{t-1}^+,\tilde\x_{t-1}, \vect 0)
    \end{align}

    \item
    Make the prediction
        \label{step:gkf:yt-}
    \begin{equation}
        \label{eq:gkf:yt-}
        \y_t^- = f_\psi(\s_t^-,\vect 0)
    \end{equation}
\end{enumerate}
Steps \ref{step:gkf:xt-enc}--\ref{step:gkf:yt-} are standard practice in state-space modeling, where $\s_{t-1}^-$ is considered as $\s_{t-1}^+$. The refinement $\s_t^+\leftarrow \s_t^-$ of the a priori state estimate $\s_t^-$ is carried out as
\begin{enumerate}[label=(\roman*)]
    \setcounter{enumi}{3}
    \item Compute the Jacobian associated with the state transition: 
    \label{step:gkf:Ft-Lt}
    \begin{align}
        \mF_{t-1} &= \nabla_{\s}f_\theta(\s_{t-1}^+,\tilde\x_{t-1},\vect 0)    \in \R^{|\Vh|\times|\Vh|}  
        \\
        \vect L_{t-1} &= \nabla_{\vetaA}f_\theta(\s_{t-1}^+,\tilde\x_{t-1},\vect 0)   \in \R^{|\Vh|\times|\Vh|^2} 
    \end{align}
    
    \item Compute the Jacobian of the readout:
    \label{step:gkf:Ht-Mt}
    \begin{align}
        \vect H_{t} &= \nabla_{\s}f_\psi(\s_{t}^-,\vect 0))    \in \R^{|\Vh|\times|\Vh|}  
        &
        \vect M_{t} &= \nabla_{\vnu}f_\psi(\s_{t}^-,\vect 0))    \in \R^{|\Vh|\times|\Vh|}  
    \end{align}
    
    \item Update the a priori error covariance: 
    \label{step:gkf:Pt-}
    \begin{align}
        \label{eq:gkf:Pt-}
        \mP_t^- &= 
            \mF_{t-1}\mP_{t-1}^+\mF_{t-1}^\top + \vect L_{t-1}\vect Q_{t-1} \vect L_{t-1}^\top 
    \end{align}
    
    \item Compute the gain matrix:
    \label{step:gkf:Kt}
    \begin{align}
        \label{eq:gkf:Kt}
        \vect K_t &= 
        \vect P_t^-\vect H_t^\top (\vect H_t \vect P_t^-\vect H_t^\top + \vect M_t\vect R_t\vect M_t^\top)^{-1}
    \end{align}

    \item Update the a posteriori state estimate:
    \label{step:gkf:st+-Pt+}
    \begin{align}
        \label{eq:gkf:st+}
        \s_t^+ &= \s_t^- + \vect K_t (\y_t - \y_t^-)
        \\
        \label{eq:gkf:Pt+}
        \mP_t^+ &= (\mathbb I - \vect K_t \vect H_t)\mP_t^-(\mathbb I - \vect K_t \vect H_t)^\top + \vect K_t \vect M_t \vect R_t\vect M_t^\top\vect K_t^\top.
    \end{align}
    
\end{enumerate}

Note that with modern libraries, such as PyTorch  \cite{paszke2019pytorch} and TensorFlow \cite{abadi2016tensorflow} along with their ecosystems \cite{fey2019fast,Cini_Torch_Spatiotemporal_2022,grattarola2021graph}, the linearization of steps \ref{step:gkf:Ft-Lt} and \ref{step:gkf:Ht-Mt} can be computed automatically in a closed form, thus allowing us to apply the proposed Graph KF to basically any deep learning architecture, including STGNNs.

\subsection{Model training and Kalman filtering}
\label{sec:GKF-training}

We train model parameters $\theta,\vartheta,$ and $\psi$ by gradient-based optimization minimizing the mean squared error (MSE) between $\y_t$ and $\y_t^-$, while the Kalman gain $\vect K_t$ and error covariance matrices $\mP_t^-$ and $\mP_t^+$ are estimated during inference. 
Note that the computations in steps \ref{step:gkf:st-} and \ref{step:gkf:Ft-Lt} 
do not involve the perturbed adjacency $\mA(\h_t)$ and noise term $\vetaA_{t-1}$ (see Assumption~\ref{a:fixed-topology}), only $\mA_\theta$; similarly, steps \ref{step:gkf:yt-} and \ref{step:gkf:Ht-Mt} do not account for the noise term $\vnu_t$. 
However, if the noise distributions have to be learned along with model parameters, above iterations can be modified to consider samples of $\vetaA_{t-1}$ and $\vnu_t$ drawn from the probabilistic model learned so far, and their probability distributions optimized by exploiting straight-through gradient estimators.

\section{Empirical validation}
\label{sec:experiments}

As our contribution is theoretical and methodological, we are just requested to validate procedures, methods, and outcomes. We leave applications to future research.
We design a set of controlled experiments to analyze the main interplaying elements: the training of a GSS model \eqref{eq:state-space-approx-assumptions}, the predictions based on the learned model (steps \ref{step:gkf:st-}--\ref{step:gkf:yt-}), and the Kalman-filter refinement (KFR) in steps \ref{step:gkf:Ft-Lt}--\ref{step:gkf:st+-Pt+}.

\subsection{Datasets}

\begin{wraptable}[8]{r}{5cm}
\vspace{-.4cm}
\caption{Parameter configuration of the GSS system models.}
\label{tab:datasets}
\small
\setlength{\tabcolsep}{4pt}
\begin{tabular}{cc|cc|cc|}
& & \multicolumn{2}{|c}{\textbf{LinGSS}} & \multicolumn{2}{|c|}{\textbf{NonLinGSS}} \\
 \hline
 $\lambda_{0}$ & $\lambda_{1}$ & 20 & 5 & 20 & 5 \\
 $\theta_\textsc{tm}$ & $\theta_\textsc{sp}$ & 0.6 & 0.3 & 0.6 & -0.3 \\
 $\psi_{0}$ & $\psi_{1}$ & -0.5 & 2.0 & -2.0 & 5.0 \\
 $\sigma_\eta$ & $\sigma_\nu$ & 0.25 & 0.12 & 0.25 & 0.12\\
 $\rho_\textsc{st}$ & $\rho_\textsc{ro}$ & id & id & $\tanh$ & $\tanh$  \\
 \hline
\end{tabular}
\end{wraptable}
We generate two GSS system models like \eqref{eq:state-space-system-model}, LinGSS and NonLinGSS, both characterized by the data-generating process detailed in next paragraphs. The specific parameters of the two GSS models are reported in Table~\ref{tab:datasets}, further details are provided as supplementary material.

\begin{figure}
\includegraphics[width=.44\textwidth]{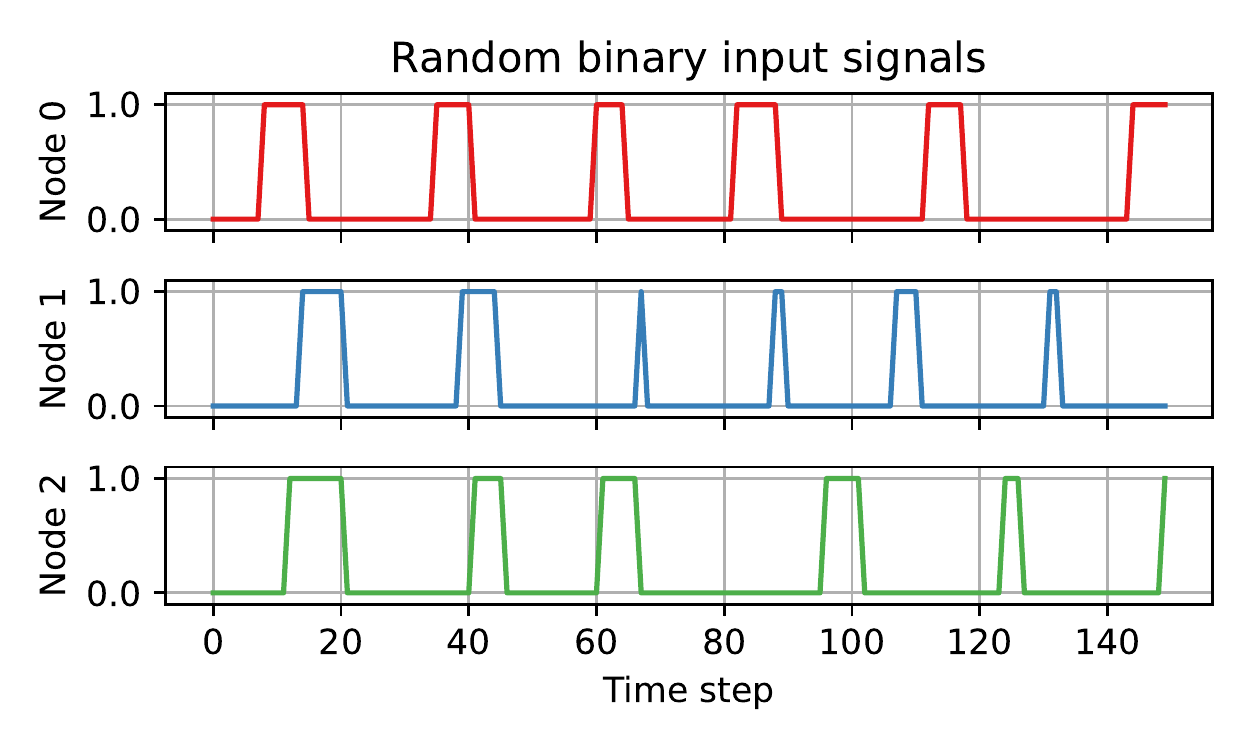}
\includegraphics[width=.44\textwidth]{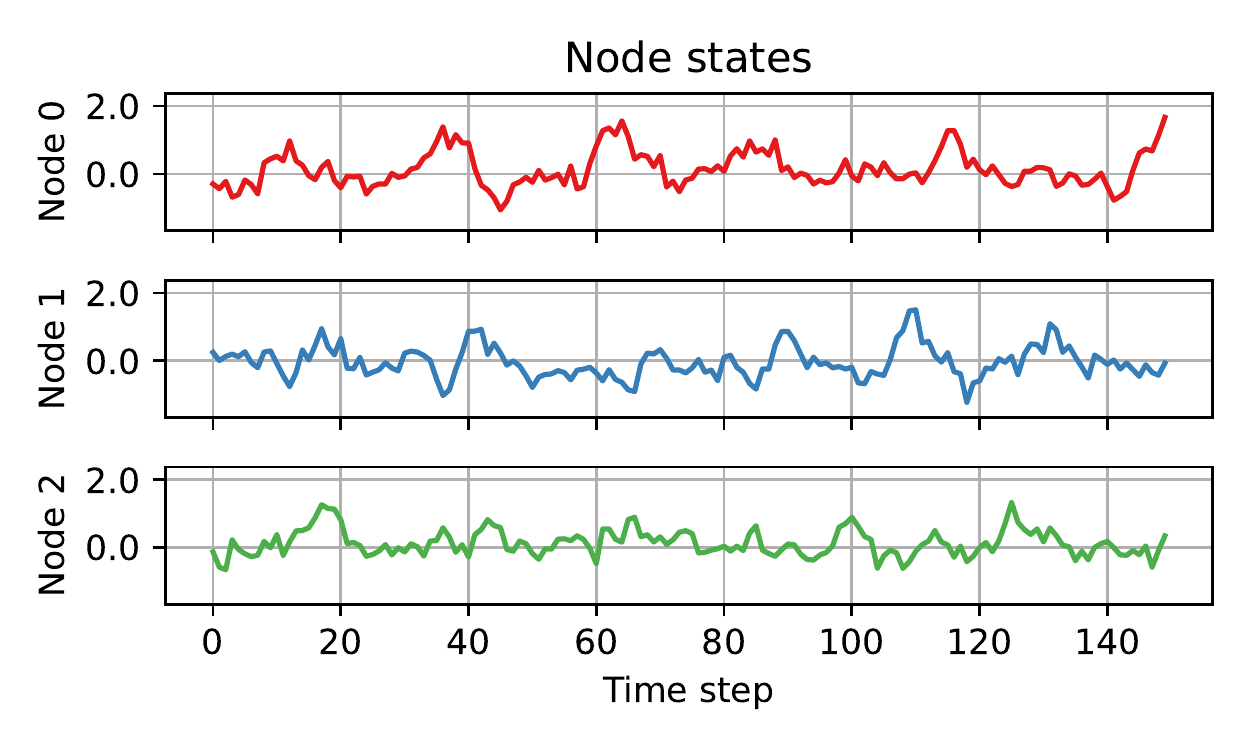}
\hfill
\includegraphics[width=.095\textwidth, trim=-0 -2cm 0 0]{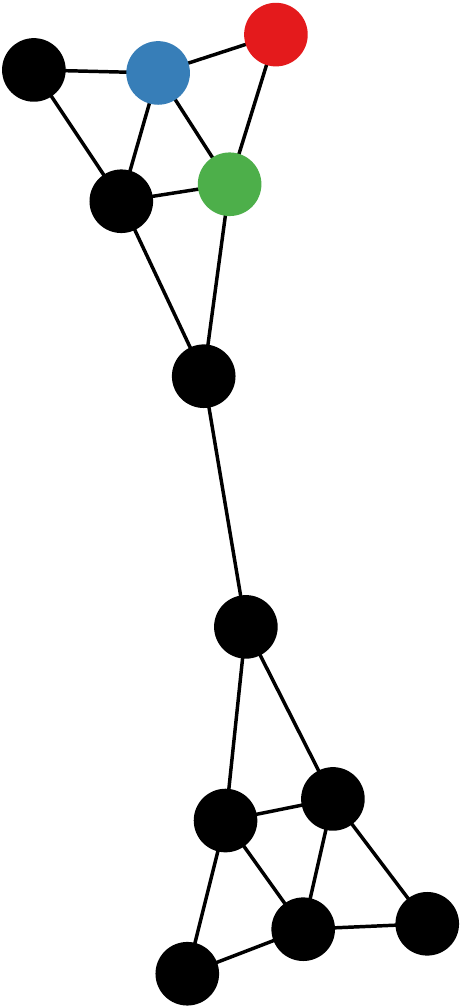}
\caption{Inputs and states of three nodes of the dataset LinGSS}
\label{fig:inputs-states-V3}
\end{figure}

\paragraph{Inputs} For each node $v\in\Vx$, input signal $s(\x_t)$ is binary and random in time duration, generated alternating between runs of 1's and 0's so that the associated run lengths are drawn from a Poisson($\lambda_1$) and a Poisson($\lambda_0$), respectively;
the left panel of Figure \ref{fig:inputs-states-V3} shows the input signals of three nodes. 
Inputs have no relational information ($E(\x_t)=\emptyset$).

\paragraph{States}
Recalling the notation $\s_t = s(\h_t)$ adopted in previous sections, states are updated according to the state-transition function
\begin{align}\label{eq:exp:gss-st}
\s_{t} &= f_\textsc{st}(\h_{t-1}, \x_{t-1}, \veta_{t-1}) 
\eqdef \rho_\textsc{st}\left(\left(\theta_\textsc{tm}\mathbb I + \theta_\textsc{sp} \bar\mA_\theta\right)\left(\s_{t-1}  + s(\x_{t-1})\right) \right)  + \veta_{t-1}
\end{align}
where $\veta_{t-1}$ is i.i.d.\ from a zero-mean Gaussian distribution with standard deviation $\sigma_\eta$, matrix $\bar\mA_\theta=\vect D^{-1/2}(\mathbb I+\mA_\theta)\vect D^{-1/2}$ is the adjacency $\mA_\theta$  normalized by diagonal matrix $\vect D$ of node degrees (self-loops included), $\Vh=\Vx$, and $\rho_\textsc{st}$ is a nonlinearity applied component-wise. Initial state $\s_0=\veta_0$ is white noise.
Figure \ref{fig:inputs-states-V3} displays the states of three sample nodes as a function of time. 

\paragraph{Outputs}
System output $\y_t$ is obtained from the same readout applied to each node-level state $\s_{t,v}$:
\begin{align}\label{eq:exp:gss-ro}
    s(\y_{t,v}) &= f_\textsc{ro}(\h_{t,v}, \vnu_{t,v})
              \eqdef \rho_\textsc{ro}(\psi_0 + \psi_1 \s_{t,v}) + \vnu_{t,v}.
\end{align}
Noise terms $\{\vnu_{t,v}\}$ are i.i.d.\ Gaussian distributed with zero as mean and $\sigma_\nu$ as standard deviation. $\rho_\textsc{ro}$ is a nonlinearity.

\subsection{Approximating family of models}

In the experiments below, we consider two types of approximating families of models \eqref{eq:state-space-approx-assumptions}.

\paragraph{Replica} The first family of models is designed to contain the state-transition and readout functions of the data-generating process in \eqref{eq:exp:gss-st} and \eqref{eq:exp:gss-ro}. In particular, the Replica model parameters are exactly the four parameters $\theta_\textsc{tm}, \theta_\textsc{sp}, \psi_0,$ and $\psi_1$.

\paragraph{STGNN} The second family of models is a generic and relatively simple STGNN that does not contain the state-transition and readout functions of the data-generating process. This family is defined by the following architecture. 
Input encoder $f_\varphi$ and the readout $f_\psi$ are 2-layer dense networks both applied node-wise.
State-transition function $f_\theta$ is composed of a message-passing layer 
\begin{align}
\s_t &= f_\theta(\s_{t-1}, f_\varphi(\x_{t-1}), \vect 0) 
\eqdef \s_{t-1} + f_\varphi(\x_{t-1}) + \tanh\left( \z \mW_\theta' + \tilde \mA_\theta \z \mW_\theta'' \right)
\end{align}
performed on $\z = \gamma_\theta(\s_{t-1} + f_\varphi(\x_{t-1}))$; $\gamma_\theta$ is a 2-layer dense network, and matrix $\tilde \mA_\theta$ is a normalized version of the adjacency matrix $\mA_\theta$ where each row adds up to 1.
All modules have 7 hidden neurons per layer and the rectified linear unit as activation function.

The models are trained to predict the expected value $\E[\y_t]$; accordingly, the MSE is considered as loss function. Parameters $\sigma_\eta, \sigma_\nu$ and the topology of the states are considered known here. Note that considering a probabilistic or dynamic topology would not change the application of the KFR. Therefore, we removed such elements from the empirical validation to focus on the KFR part of the proposed Graph KF.
Further experimental details are given in the supplementary material.

\subsection{Positive effect of the Graph KF refinement}

In the first experiment, we study the improvement brought by the KFR assuming to know the system model, i.e., the Replica model with parameters identical to those in Table~\ref{tab:datasets}. We compare the prediction error $\lVert\y_t^- - \y_t\rVert_2^2$ when $\y_t^-=f_\textsc{ro}(\hat\h_t)$ has been computed from the following different states $\hat\h_t$: 
\begin{description}[wide, labelwidth=!, labelindent=1pt]
    \item[(w/o KFR) ] $\hat\h_t = \h_t^-$ is the state estimate produced from the past states $\{\h_i:i<t\}$ and inputs $\{\x_i: i\le t\}$ by performing steps \ref{step:gkf:xt-enc}--\ref{step:gkf:yt-} with $f_\textsc{st}$ as state transition.
    \item[(w/ KFR) ] $\hat\h_t = \h_t^+$ is the state estimate produced from the past states $\{\h_i:i<t\}$ and inputs $\{\x_i: i\le t\}$ by performing steps \ref{step:gkf:xt-enc}--\ref{step:gkf:st+-Pt+} with $f_\textsc{st}$ as state transition, thus applying also the KFR.
    \item[(Exp) ] $\hat\h_t = \E[\h_t]$ is the expected value of the state computed from \eqref{eq:exp:gss-st} without the noise term.
    \item[(GT) ] $\hat\h_t = \h_t$ is the true system state in \eqref{eq:exp:gss-st} affected by noise, thus serving as ground truth.
\end{description}
Results are reported in Table~\ref{tab:replica-models}.

\paragraph{Results w/o KFR} On LinGSS dataset, Replica GT relies on the true state and its performance matches the readout noise $\sigma_\nu^2=(0.12)^2=0.0144$, whereas the Replica Exp performance is close to the target value $\sigma_\nu^2 + (\sigma_\eta\psi_1)^2 =0.2644$.
Note that although the state transition and readout are exactly those that generated the data, the noise on the state transition has a negative impact on the prediction performance. 
Therefore, and as expected, Replica model without the KFR performs worse than both baselines Replica Exp and Replica GT.

\paragraph{Results w/ KFR} Conversely, KFR allows Replica model to approach the baseline performance of Replica Exp, while the performance of the reference model Replica GT cannot be achieved without information about the random realization of $\veta_t$ noise. 
Table~\ref{tab:replica-models} displays also the following relative prediction improvement (RPI)
\begin{equation}
\frac{\lVert\y_t^+-\y_t\rVert_2^2 - \lVert\y_t^--\y_t\rVert_2^2}{\lVert\y_t^--\y_t\rVert_2^2}
\end{equation}
to further compare a priori and a posteriori estimates. The RPI values in Table~\ref{tab:replica-models} show that the prediction error is reduced to almost zero (equivalent to RPI of -100\%), reassuring of the correct functioning of the KFR procedure. We should note that $\y_t^+$ here is receiving $\y_t$ as feedback to update the state estimate, however, we stress that for computing the prediction error in the first column of the table, the target $\y_t$ is \emph{not} seen by the predictive model. For the same reason, observe also that despite the RPI values, applying the KFR to Replica GT and Replica Exp does not bring further down the MSE, as they are already optimal in their respective senses.

\paragraph{Results on NonLinGSS} Similar results are observed for NonLinGSS dataset, too, where Replica model with KFR improves over Replica model without KFR, and whose performance approaches that of Replica Exp; we comment that in this nonlinear setting, the performance of Replica Ext is not expressed as $\sigma_\nu^2 + (\sigma_\eta\psi_1)^2$, as it depends on the nonlinearity $\rho_\textsc{ro}$, as well.

\begin{table}
\centering
\caption{Performance of Replica model with and without the KFR. The prediction error is a single value as the model parameters are predefined. The RPI, reported in the form ``mean\textsubscript{$\pm$std}'' is estimated over the test mini-batches.}
\label{tab:replica-models}
\small
\begin{tabular}{l|cc|c|cc|c|}
 & \multicolumn{3}{c|}{LinGSS} & \multicolumn{3}{c|}{NonLinGSS} \\
 & \multicolumn{2}{c|}{\textbf{Pred.\ Err.} {\footnotesize(MSE)}} & \textbf{RPI} & \multicolumn{2}{c|}{\textbf{Pred.\ Err.} {\footnotesize(MSE)}} & \textbf{RPI}\\ \textbf{Model} &\footnotesize w/o KFR &\footnotesize w/ KFR & \footnotesize (MSE\%)&\footnotesize w/o KFR &\footnotesize w/ KFR &\footnotesize (MSE\%)\\
\hline
 Replica     & 0.384 & 0.271 & -98.6\textsubscript{$\pm$1.2} & 0.483 & 0.359 & -88.7\textsubscript{$\pm$1.6}\\
 Replica Exp & 0.267 & 0.267 & -98.8\textsubscript{$\pm$0.6} & 0.349 & 0.349 & -85.6\textsubscript{$\pm$2.2}\\
 Replica GT  & 0.014 & 0.014 & -98.9\textsubscript{$\pm$0.0} & 0.014 & 0.014 & -62.3\textsubscript{$\pm$4.5}\\ 
\hline
\end{tabular}
\end{table}

\subsection{Graph KF refinement on trained models}

The second set of experiments concerns performance improvements when using approximating models. In this problem setup, we expect that the trained models fit well the data-generating process. However, we do not expect them to identify (match exactly) the state transition and readout of the system model. 

Table~\ref{tab:trained-models} shows that all considered models benefit from the KFR.
In particular, we observe that the Replica model trained on LinGSS matches the performance of Table~\ref{tab:replica-models}, where the model parameters were set equal to those that generated the data (Table~\ref{tab:datasets}). Moreover, inspecting the learned parameters, we see they get close to the ground truth reported in Table~\ref{tab:datasets}. 
Interestingly, the parameters learned by Replica on NonLinGSS differ from the ground truth ($\sim$10\% off) and enable better predictions than those of Table~\ref{tab:replica-models}.
Finally, the STGNN models (whose family does not contain the system model) do not performed as good as Replica ones, especially when comparing the refined versions, as confirmed also by the RPI. In spite of that, we stress that STGNN benefits from the KFR and, in fact, displays an RPI smaller than zero and achieves a better MSE than all the other models without KFR.

\begin{table}
\caption{Performance of trained models with and without the KFR. The prediction error is averaged over 10 runs whereas the RPI is estimated over the test mini-batches and the 10 runs. Results are reported in the format ``mean\textsubscript{$\pm$std}''.}
\label{tab:trained-models}
\small
\centering
\begin{tabular}{l|cc|c|cc|c|}
 & \multicolumn{3}{c|}{LinGSS}  & \multicolumn{3}{c|}{NonLinGSS}  \\
 & \multicolumn{2}{c|}{\textbf{Pred.\ Err.} {\footnotesize (MSE)}} & \textbf{RPI} & \multicolumn{2}{c|}{\textbf{Pred.\ Err.} {\footnotesize(MSE)}} & \textbf{RPI}   \\
 \textbf{Model} &\footnotesize w/o KFR & \footnotesize w/ KFR & \footnotesize (MSE\%) & \footnotesize w/o KFR & \footnotesize w/ KFR & \footnotesize (MSE\%)\\
 \hline
 Replica & 0.384\textsubscript{$\pm$0.000} & 0.271\textsubscript{$\pm$0.000} & -98.6\textsubscript{$\pm$1.1\ \;} 
    & 0.429\textsubscript{$\pm$0.000} & 0.327\textsubscript{$\pm$0.000} & -94.0\textsubscript{$\pm$0.8} \\
 STGNN & 0.389\textsubscript{$\pm$0.001} & 0.336\textsubscript{$\pm$0.019} & -34.2\textsubscript{$\pm$15.9} 
     & 0.434\textsubscript{$\pm$0.001} & 0.407\textsubscript{$\pm$0.010} & -13.5\textsubscript{$\pm$7.7 }\\
 \hline
\end{tabular}
\end{table}

\section{Conclusions}
\label{sec:conclusions}
This paper extends for the first time the Kalman and extended Kalman filters to graphs, i.e., inputs, states, and outputs are attributed graphs of variable topology. The contribution is theoretical and, as such, an experimental section is meant only at validating the developments.
Hypotheses are made to facilitate the reading and derivation thanks to a lighter notational setup; their relaxation is the subject of future research that, however, does not change the spirit of what is here proposed.

\subsection*{Acknowledgement}
This work was supported by the Swiss National Science Foundation project FNS 204061: \emph{High-Order Relations and Dynamics in Graph Neural Networks}.

\bibliographystyle{plainnat}
\bibliography{biblio}



\appendix


\section{Hardware and software}

The code for the empirical evaluation of the proposed method has been developed in Python relying on the following open-source libraries: PyTorch~\cite{paszke2019pytorch}, PyTorch Geometric~\cite{fey2019fast}, Torch Spatiotemporal~\cite{Cini_Torch_Spatiotemporal_2022}, PyTorch Lightning~\cite{Falcon_PyTorch_Lightning_2019} and NumPy~\cite{harris2020array}. The experiments were run on machines equipped with AMD EPYC 7513 processors and NVIDIA RTX A5000 GPUs.

\section{Model training}
The models are trained to minimize the mean squared error (MSE) with Adam optimizer for 100 epochs and learning rate of 0.01. 70\% of the data is used for training, 10\% for validation, and 20\% for testing.
The batch size is set to 32 and the predictions are made from a window size of 12 time steps. Early stopping on the best validation MSE is applied with a 10-epoch patience. Training a model typically takes less than 10 minutes for Replica models and about 20 minutes for the STGNNs. 

\section{Dataset visualization}

\begin{figure}
\centering
\includegraphics[width=.48\textwidth]{img/LinSTSSv3_inputs.pdf}
\includegraphics[width=.48\textwidth]{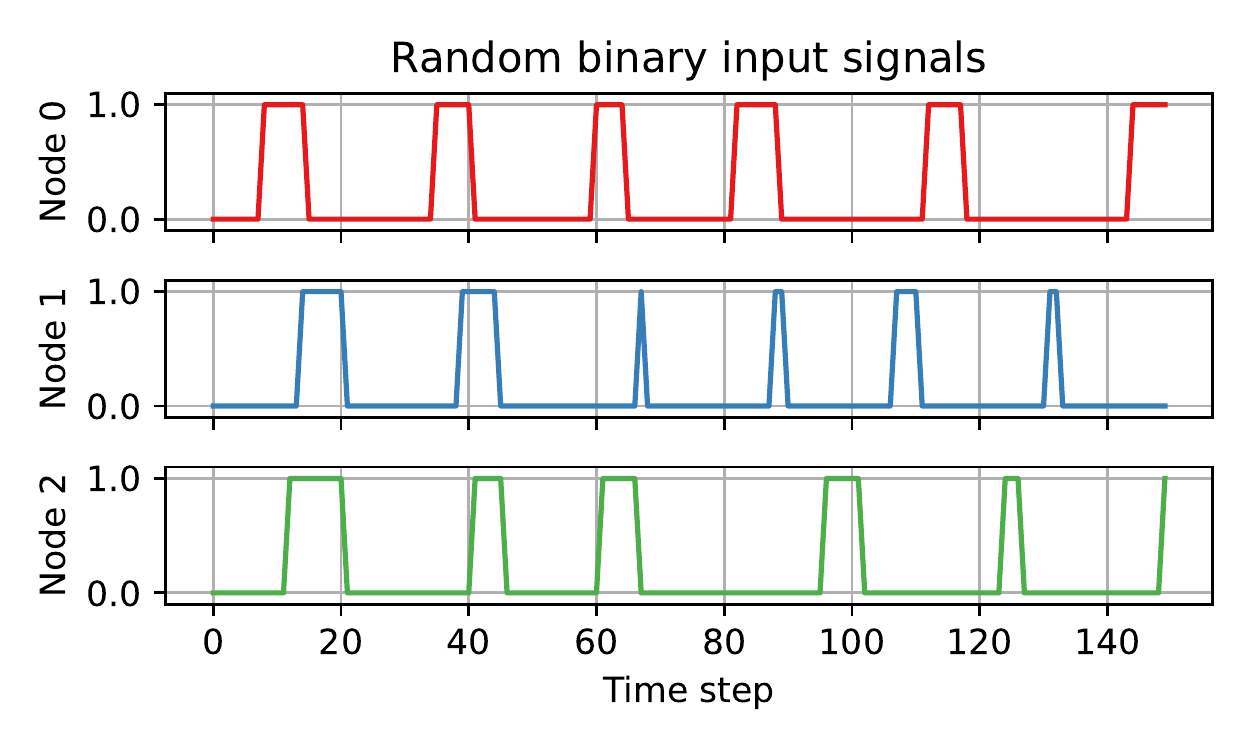}
\\
\includegraphics[width=.48\textwidth]{img/LinSTSSv3_states.pdf}
\includegraphics[width=.48\textwidth]{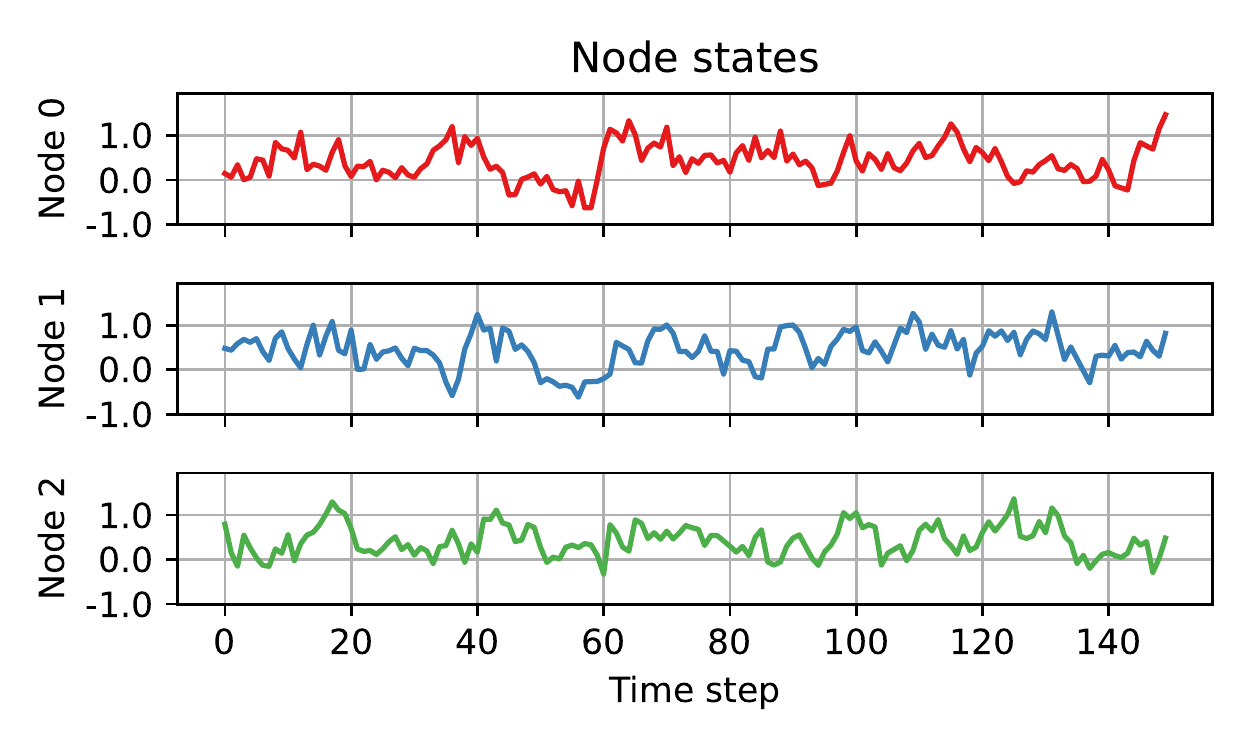}
\\
\includegraphics[width=.48\textwidth]{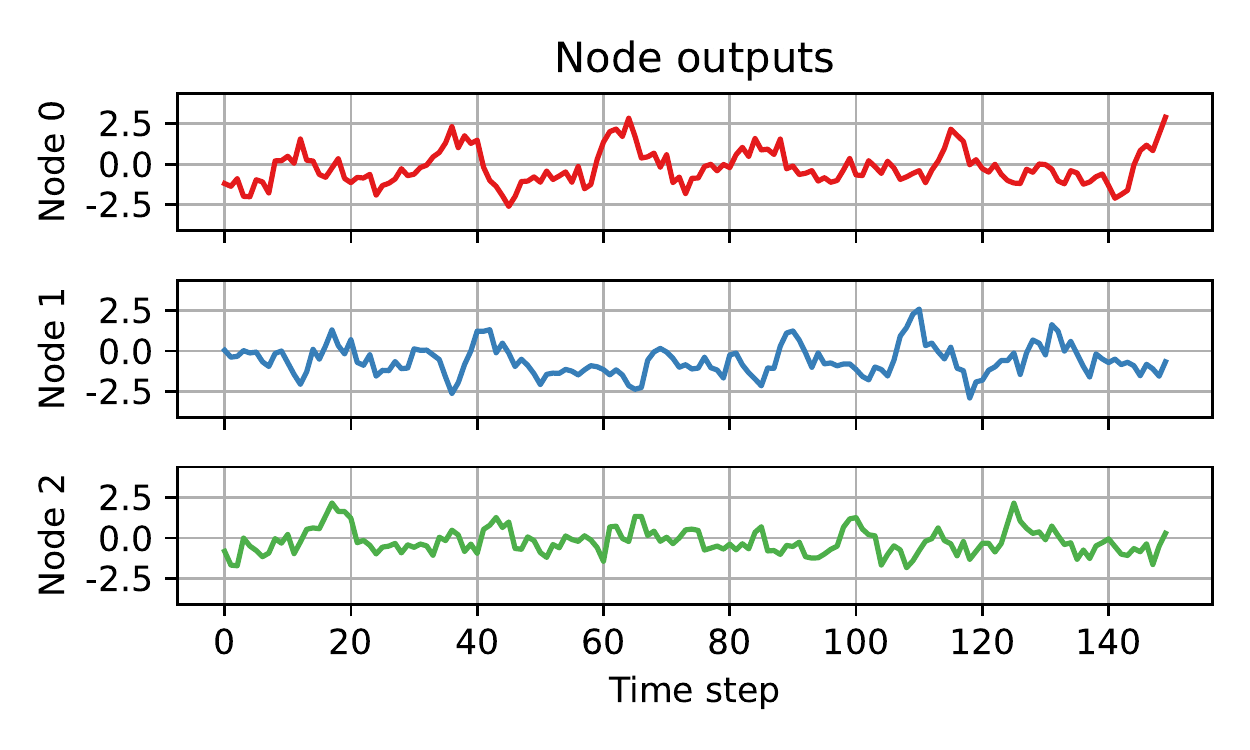}
\includegraphics[width=.48\textwidth]{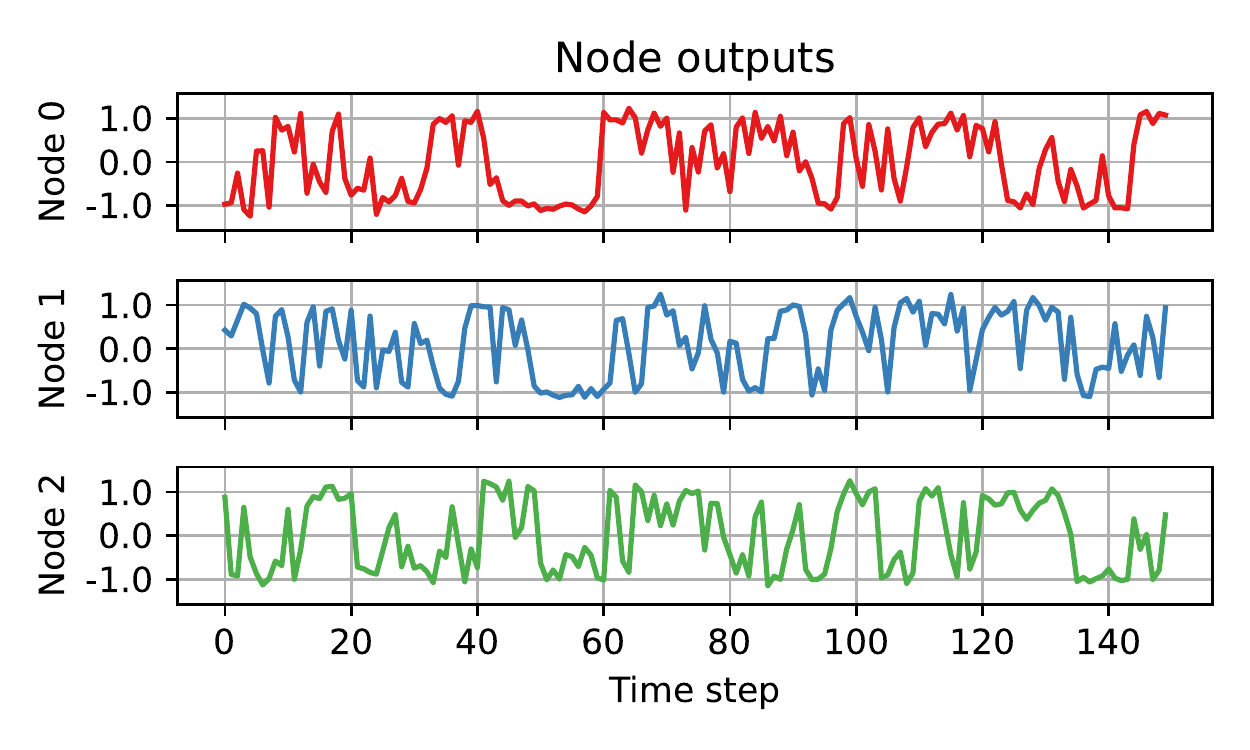}
\\
\includegraphics[width=.16\textwidth, trim=0 0cm 0 0, angle=90,origin=c]{img/graph-crop.pdf}
\caption{Inputs, states, and outputs of three nodes from LinGSS dataset (left) and NonLinGSS (right).}
\label{fig:app:inputs-states-outputs}
\end{figure}

Figure~\ref{fig:app:inputs-states-outputs} shows examples of node-level inputs, states, and outputs of three nodes in LinGSS and NonLinGSS. The considered state graphs have 12 nodes connected as shows at the bottom of the figure.

\end{document}